%% file: main.tex
\documentclass[10pt,twocolumn,letterpaper]{article}

\usepackage{cvpr}              

\input{preamble}

\definecolor{cvprblue}{rgb}{0.21,0.49,0.74}
\usepackage[pagebackref,breaklinks,colorlinks,allcolors=cvprblue]{hyperref}
\usepackage{enumitem}

\setlist[itemize]{itemsep=0em, topsep=0em} 
\usepackage{bm}
\usepackage{amsmath}
\usepackage{amssymb}
\usepackage{multirow}
\usepackage{gensymb}
\usepackage{makecell} 
\def\paperID{8471} 
\def\confName{ICCV}
\def\confYear{2025}

\title{MonoMobility: Zero-Shot 3D Mobility Analysis from Monocular Videos}

\author{
Hongyi Zhou$^{1}$~~~~~Yulan Guo$^{2}$~~~~~Xiaogang Wang$^{3}$\footnotemark[1]~~~~~Kai Xu$^{1}$\thanks{Xiaogang Wang and Kai Xu are corresponding authors.}
\vspace{4pt}\\
$^1$National University of Defense Technology\\
$^2$Sun Yat-sen University  \\
$^3$College of Computer and Information Science, Southwest University
}

\begin{document}
\maketitle
\input{sec/0_abstract}    
\input{sec/1_intro}
\input{sec/2_related_work}
\input{sec/3_method}
\input{sec/4_experiment}
\input{sec/5_conclusion}
\clearpage
\input{sec/7_acknowledge}
{
    \small
    \bibliographystyle{ieeenat_fullname} 
    \bibliography{main}
}


\end{document}

%% file: preamble.tex
\usepackage[utf8]{inputenc}


%% file: sec/0_abstract.tex
\begin{abstract}
Accurately analyzing the motion parts and their motion attributes in dynamic environments is crucial for advancing key areas such as embodied intelligence. Addressing the limitations of existing methods that rely on dense multi-view images or detailed part-level annotations, we propose an innovative framework that can analyze  3D mobility from monocular videos in a zero-shot manner. This framework can precisely parse motion parts and motion attributes only using a monocular video, completely eliminating the need for annotated training data. Specifically, our method first constructs the scene geometry and roughly analyzes the motion parts and their initial motion attributes combining depth estimation,  optical flow analysis and point cloud registration method, then employs 2D Gaussian splatting for scene representation. Building on this, we introduce an end-to-end dynamic scene optimization algorithm specifically designed for articulated objects, refining the initial analysis results to ensure the system can handle ‘rotation’, ‘translation’, and even complex movements (‘rotation+translation’), demonstrating high flexibility and versatility. To validate the robustness and wide applicability of our method, we created a comprehensive dataset comprising both simulated and real-world scenarios. Experimental results show that our framework can effectively analyze articulated object motions in an annotation-free manner, showcasing its significant potential in future embodied intelligence applications. The project page is at: \url{https://monomobility.github.io/MonoMobility}.
\end{abstract}

%% file: sec/1_intro.tex
\vspace{-10pt}
\section{Introduction}
\label{sec:intro}

\begin{figure}[t]
  \centering
  \includegraphics[width=1\linewidth]{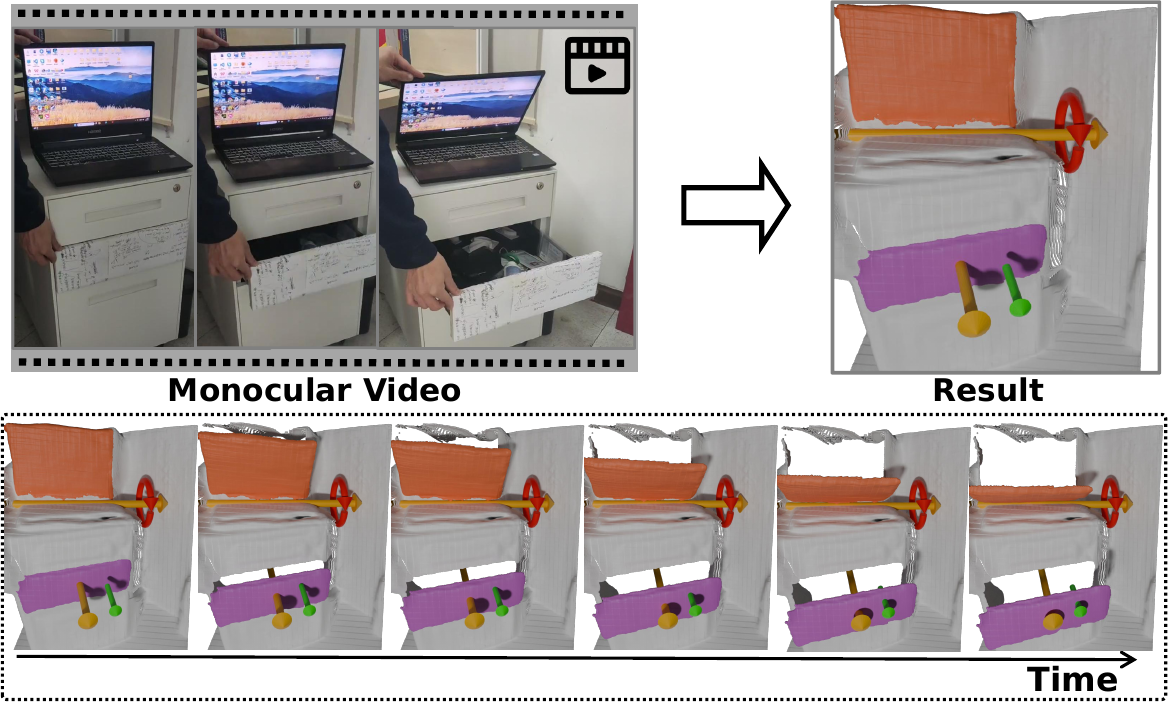}

   \caption{Our method could parse the motion parts of articulated objects and their corresponding motion attributes from a monocular video (top left). Under the constraints of motion attributes, the parsed parts can be driven to move realistically (bottom).}
   \label{fig:teaser}
   \vspace{-10pt}
\end{figure}

Accurately parsing the motion parts and their motion attributes of articulated objects within dynamic environments is a critical cornerstone for embodied intelligence and related disciplines. For instance, our daily surroundings are populated with numerous articulated objects, including drawers, swivel chairs, staplers, and pump bottles. Efficient robot interaction with these items, whether it involves pulling out drawers or pressing pump heads, hinges on a precise grasp of the objects' structure and movement characteristics within the scene, particularly the identification of their motion parts and motion attributes. Consequently, the ability to accurately parsing the motion parts and their motion attributes of articulated objects in dynamic settings is not only foundational but also of paramount importance.

Currently, substantial research efforts have been directed towards the analysis of articulated objects \cite{jiang2022opd,sun2024opdmulti,collins20232,shi2021self,shi2022p,geng2023gapartnet,deng2024banana,liu2022toward,li2020category,zeng2021visual,liu2023semi,che2024op,fu2024capt,wang2019shape2motion,yan2019rpm,qian2022understanding,liu2023paris,patil2023rosi,weng2024neural,swaminathan2024leia,mandi2024real2code,shi2024articulated,jiang2022ditto,wang2024sm,du2023learning,kawana2024detection,song2024reacto,haresh2022articulated,mu2021sdf,li2017grass,gu2025artiscene}. Shape2Motion \cite{wang2019shape2motion} predicts motion parts and their attributes from 3D point clouds of articulated objects. However, this data-driven method often struggles with unseen data, leading to suboptimal performance. PARIS \cite{liu2023paris} and Weng et al. \cite{weng2024neural} employ dense RGB(D) multi-view images of the initial and final states to reconstruct both static and dynamic parts of articulated objects, as well as analyze the motion attributes of these parts. Nevertheless, obtaining dense multi-view images with precise pose information poses significant challenges, making it difficult to apply these techniques in real-world settings with limited fields of view. Furthermore, these methods primarily operate at the object level, which hinders their scalability to complex scenarios. Recently, OPDMulti \cite{sun2024opdmulti} is capable of recognizing multiple articulated objects and predicting their motion attributes from a single RGB image. Despite this advancement, OPDMulti's capabilities are confined to openable objects and do not extend to irregularly structured articulated objects, such as staplers.

Considering that robots usually are equipped with RGB cameras and operate within a limited field of view, we use monocular video as the input condition. Meanwhile, in the operating environment, there is often complex background. Therefore, summarizing the above practical requirements, we describe this task, which is to analyze the motion parts of articulated objects and their corresponding motion attributes(motion axis and motion type) at scene level from captured monocular videos, as shown in Figure \ref{fig:teaser}.

For analyzing dynamic articulated objects from monocular videos, our key insights are: (1) Captured videos of dynamic articulated objects are a treasure trove of 3D movement information. By leveraging this data through off-the-shelf methods to obtain potential geometric and motion information, we can reconstruct the scene and conduct an initial analysis of articulated objects. (2) The correctness of identified motion parts and the accuracy of motion parameter analysis improves as the estimated dynamic process aligns more closely with the ground-truth motion pattern. Therefore, we can refine and improve the analysis results by optimizing the dynamic articulated scene.

Specifically, regarding the input video, we first estimate the camera poses, extract depth maps and optical flow using off-the-shelf methods ~\cite{yang2024depth, he2025distill, le2024dense, teed2020raft, leroy2024grounding}, followed by the segmentation of flow map~\cite{xie2024moving}. This step yields a segmented point cloud for the motion parts, which may contain outliers of static region, serving as the basis for initializing the 2D Gaussians. During initialization, we assume all motion parts undergo composite motion(rotation+translation), and estimate the motion axis by utilizing the transformation matrix between the first and last frame point clouds of each motion part. The optimization phase identifies the actual motion parts and refines their motion parameters by iteratively sampling pairs of frames, transforming the Gaussians of the motion parts to the frames based on the current estimates of translation and rotation, which calculated from the  motion axis and motion quantity, and optimizing the scene states through a combination of rendering, normal, and motion losses. The rendering loss ensures visual consistency, the normal loss improves geometric accuracy, and the motion loss ensures the motion pattern consistency between the estimated dynamics and the point cloud transformations. After sufficient iterations, we remove the misidentified motion parts and classify the motion type of actual motion parts based on the translational and rotational movements, applying the realistic transformations thereafter. Our approach allows for accurate analysis of dynamic articulated objects from a monocular video, capturing the realistic motions of articulated objects effectively.

In summary, our contributions are as follows:
\begin{itemize}
\item We propose a novel zero-shot framework for analyzing dynamic articulated objects from monocular videos, effectively capturing motion parts and motion attributes.
\item We develop a self-supervised dynamic scene optimization approach capable of refining motion attributes, including motion type and parameters.
\item We construct a dataset featuring dynamic scenes with articulated objects in both synthetic and real-world settings, designed to rigorously evaluate related techniques.
\end{itemize}

%% file: sec/2_related_work.tex
\begin{figure*}[t]
  \centering
  \includegraphics[width=1\linewidth]{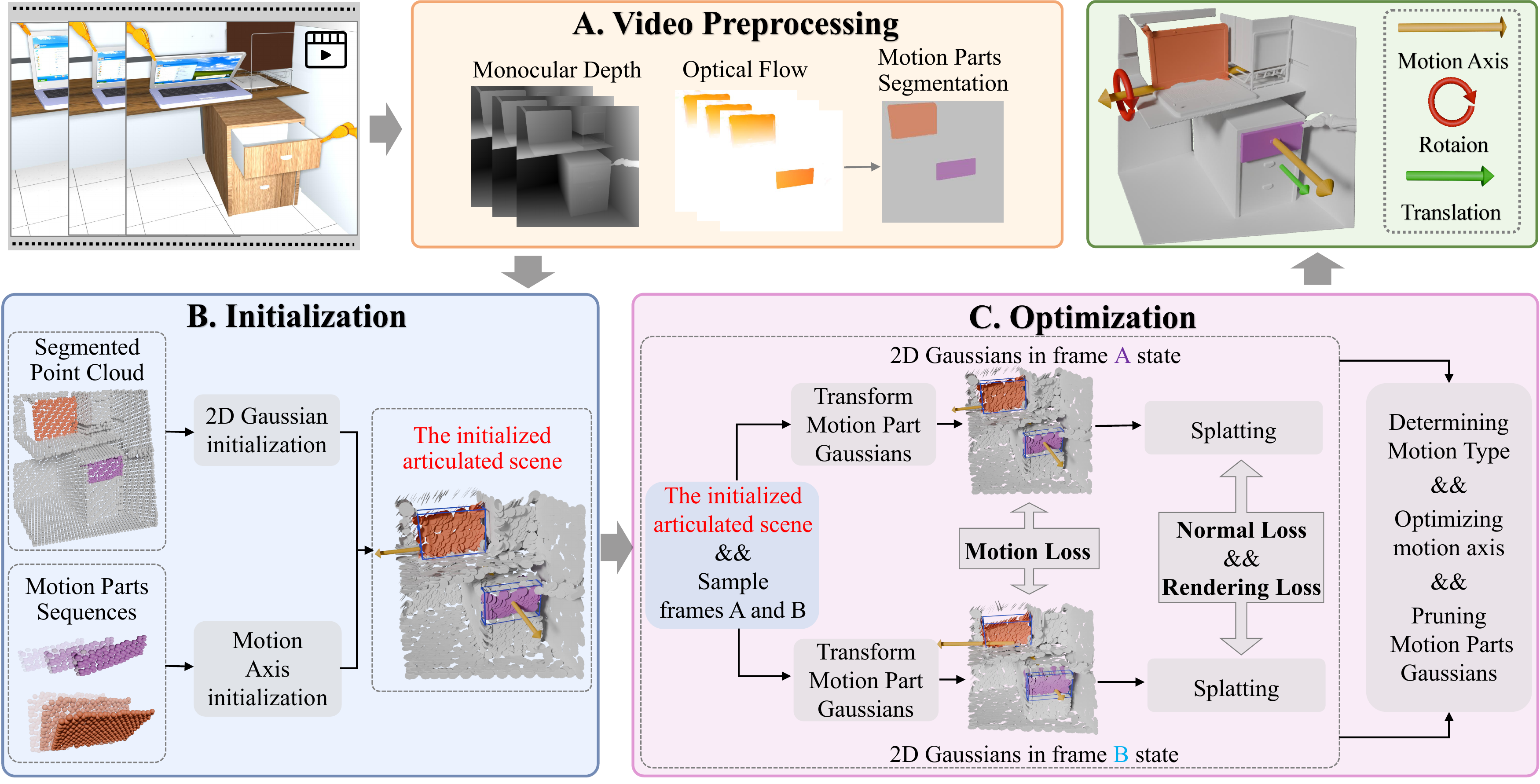}

   \caption{Overview of our method: Given a monocular video of dynamic articulated objects, we analyze the motion parts and their motion attributes. Initially, we utilize off-the-shelf methods to obtain the segmented point clouds of the static scene and the motion parts sequences over time. Subsequently, we initialize the 2D Gaussians of articulated scene, incorporating motion parts labels and motion axes. Finally, we employ end-to-end dynamic scene optimization method (specially designed for articulated objects) to optimize motion axes, determine motion type, and prune motion parts.}
   \label{fig:pipeline}
   \vspace{-10pt}
\end{figure*}

\vspace{-3pt}
\section{Related Work}
\label{sec:Related Work}
\vspace{-5pt}
\noindent\textbf{\textbf{Data-Driven Articulated Object Analysis.}} A significant body of research~\cite{jiang2022opd,sun2024opdmulti,collins20232,shi2021self,shi2022p,geng2023gapartnet,deng2024banana,liu2022toward,li2020category,zeng2021visual,liu2023semi,che2024op,fu2024capt,wang2019shape2motion,yan2019rpm,qian2022understanding} employs deep learning on large-scale annotated datasets to identify motion parts and estimate motion attributes. Shape2Motion~\cite{wang2019shape2motion} predicts motion parts and attributes from object point clouds, while OP-Align~\cite{che2024op} requires segmentation information from RGB-D frames. Compared to 3D point clouds and depth maps, RGB images are more accessible. OPD~\cite{jiang2022opd} and OPDMulti~\cite{sun2024opdmulti} predict motion parts and attributes from RGB images but struggle with irregular articulated objects like staplers.

\noindent\textbf{\textbf{Reconstruction and Analysis of Articulated Objects.}} Methods~\cite{liu2023paris,patil2023rosi,weng2024neural,swaminathan2024leia,mandi2024real2code,shi2024articulated,jiang2022ditto,wang2024sm,du2023learning,kawana2024detection,song2024reacto,haresh2022articulated,mu2021sdf,liu2022efficient} reconstruct articulated objects using 3D or multi-view data. A-SDF~\cite{mu2021sdf} pioneers neural implicit representations for articulated object reconstruction, requiring 3D supervision. Ditto~\cite{jiang2022ditto} reconstructs articulated structures and analyzes motion parameters from two-state object point clouds. PARIS~\cite{liu2023paris} uses multi-view images of initial and final states to reconstruct static and motion parts, analyzing motion attributes. Weng et al.~\cite{weng2024neural} leverages RGB-D images for enhanced geometry and motion understanding, while Shi et al.~\cite{shi2024articulated} employs generative adversarial networks to reduce reliance on temporal information. These methods' dependence on complete 3D data or dense multi-view images limits their practicality in environments with limited fields of view.

\noindent\textbf{\textbf{3D Gaussian-Based Dynamic Scene Reconstruction.}} 3D Gaussian-based approaches~\cite{wu20244d,stearns2024dynamic,wang2024shape,gao2025curveawaregaussiansplatting3d,gao2025selfsupervisedlearninghybridpartaware} achieve state-of-the-art performance in 3D/4D reconstruction. 4D Gaussian Splatting~\cite{wu20244d} predicts Gaussian attribute changes over time using a six-plane dynamic module. Subsequent works~\cite{stearns2024dynamic,wang2024shape} explore single-view Gaussian dynamic reconstruction. DGMarbles~\cite{stearns2024dynamic} and Shape of Motion~\cite{wang2024shape} leverage advanced data-driven techniques to obtain geometric and motion priors. Unlike these works, which treat each Gaussian independently, our method unifies motion part Gaussians to capture realistic motion patterns.

%% file: sec/3_method.tex
\vspace{-5pt}
\section{Problem Statement}
\label{sec:problem statement}
\vspace{-3pt}
For a monocular video capturing the motion of articulated objects, represented as a sequence of image frames $I=\{I_i\in{\mathbb{R}^{H\times W\times 3}}\}_{i=1}^N$. Our goal is to identify the motion parts $\mathcal{M}=\{M^x\}_{x=1}^X$ of articulated objects and analyze their motion attributes $\mathcal{A}=\{A^x\}_{x=1}^X$, where $A^x=<t_x, p_x>$, $t_x$ and $p_x$ represent motion type and motion parameters respectively (see Fig. \ref{fig:teaser}). The motion types comprise translation (\textbf{T}), rotation (\textbf{R}), and simultaneous rotation with translation (\textbf{RT}). The motion parameters $p$ represents motion axis, including axis direction $\vec{r}\in{\mathbb{R}^3}$ and axis position $q\in{\mathbb{R}^3}$. The translational motion part moves along the axis direction,  it is only related to the direction of the axis $\vec{r}$. For the motion part of the other two motion types, rotation occurs around the axis, therefore, in addition to the axis direction, the axis position $q$ must also be considered.

\vspace{-3pt}
\section{Method} 
\label{sec:method} 
\vspace{-3pt}
The dynamic scene of articulated objects is a special type of dynamic scene, where the motion parts move realistically under the constraints of motion attributes. We utilize the optimization method of dynamic scenes to parse the motion parts and motion attributes of these articulated objects. In this section, we adopt the 2D Gaussian splatting~\cite{huang20242d} as the scene representation (Section \ref{sec:Scene Representation}). To achieve dynamic scene reconstruction from monocular videos, we first use off-the-shelf methods~\cite{yang2024depth, he2025distill, luo2024flowdiffuser, le2024dense,leroy2024grounding, xie2024moving, wang2024dust3r} to estimate camera poses, extract depth maps and optical flow, then segment the optical flow map. This step constructs the scene geometry using point clouds extracted from the depth maps and divides the scene into static region and motion parts, some of which may be incorrectly identified and actually belong to the static region (Section \ref{sec:Video Frames Preprocessing}). Subsequently, We use point cloud data to initialize static region and motion parts as 2D Gaussian representation, and initialize motion attributes to construct the entire dynamic scene(Section \ref{sec:Initialization}), finally retain the actual motion parts and refine their motion attributes by optimizing the dynamic scene (Section \ref{sec:Optimization}). Figure \ref{fig:pipeline} illustrates the overall structure of our method.

\begin{figure*}[t]
  \centering
  \includegraphics[width=1\linewidth]{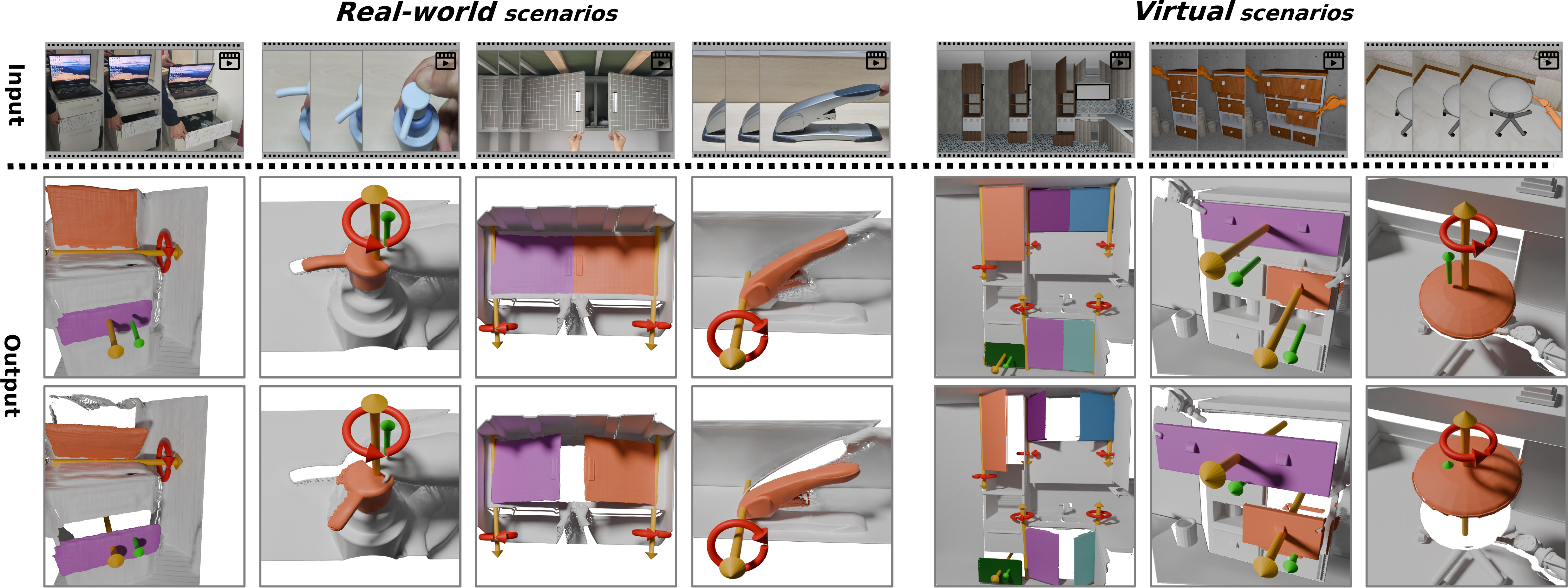}

   \caption{Qualitative results. Our method can handle real-world and virtual datas. With the predicted motion axis, the motion parts are able to execute realistic motions.}
   \label{fig:real_world}
   \vspace{-8pt}
\end{figure*}
\subsection{Scene Representation}
\label{sec:Scene Representation}
\noindent\textbf{\textbf{Preliminary: 2D Gaussian Splatting.}}
In this work, we adopt the 2D Gaussians splatting~\cite{huang20242d} for scene representation because it supports fast, differentiable rendering and its explicit representation facilitates the integration of motion parts with motion attributes.

2D Gaussians can be parameterized as $\mathcal{G}=(\bm{x}, \bm{r}, \bm{s_{uv}}, \bm{\sigma}, \bm{c)}$, where $\bm{x}$ represents the position; $\bm{r}$ denotes rotation; $\bm{s_{uv}}$ controls the variance of the Gaussian distribution; $\bm{\sigma}$ indicates the opacity, and $\bm{c}$ represents the viewpoint-dependent color of the 2D Gaussian, which is described by spherical harmonics coefficients.

During rendering, the image is first divided into 16$\times$16 tiles, with each tile being processed in parallel to identify the 2D Gaussians that intersect with pixel rays and calculate the intersection coordinates $\bm{u}=(u,v)$ in the $uv$ plane of the 2D Gaussian. The Gaussian value is then computed: 
\vspace{-5pt}
\begin{equation}
  {G}(\bm{u})=\exp\left(-\frac{u^2+v^2}{2}\right)
\end{equation}

Finally, based on the depth value of each Gaussian point in the camera coordinate system, $\alpha-blending$ is performed from front to back to obtain the color for each pixel: 
\vspace{-5pt}
\begin{equation}
c=\sum_{i=1}c_i\sigma_i{G}_i(\bm{u}_i)\prod_{j=1}^{i-1}(1-\sigma_j {G}_j(\bm{u}_j))
\end{equation}

\subsection{Video Preprocessing}
\label{sec:Video Frames Preprocessing}
For the input video $\mathcal{I}=\{I_i\in{\mathbb{R}^{H\times W\times 3}}\}_{i=1}^N$, we first employ depth estimation methods~\cite{yang2024depth, he2025distill, hu2024metric3d} to obtain a sequence of depth maps $\mathcal{D}=\{D_i\in{\mathbb{R}^{H\times W}}\}_{i=1}^N$ and use MASt3R~\cite{leroy2024grounding} to estimate the corresponding camera poses $\mathcal{P}o=\{Po_i\in{\mathbb{R}^{4\times 4}}\}_{i=1}^N$. Then, combining the camera intrinsic parameters and the camera poses $\mathcal{P}o$, we extract a sequence of point clouds $\mathcal{P}c=\{Pc_i\in{\mathbb{R}^{H\times W\times 3}}\}_{i=1}^N$ in the world coordinate space from the depth map sequence $\mathcal{D}$. In order to better analyze the motion of articulated objects in the monocular video, we utilize optical flow methods~\cite{le2024dense, teed2020raft, luo2024flowdiffuser} to acquire the pixel trajectory sequence $\mathcal{F}=\{F_i\in{\mathbb{R}^{H\times W\times 2}}\}_{i=1}^N$. Subsequently, we apply FlowI-SAM~\cite{xie2024moving} to segment $\mathcal{F}$, resulting in instance-level segmentation masks of motion parts: $\mathcal{M}=\{M_i \in \{0, 1, \dots, K\}^{H \times W}\}_{i=1}^N$, where 0 represents the static region, and $1-K$ denotes the segmented motion parts (with $K \geq X$, where $K$ is the number of segmented motion parts and $X$ is the number of actual motion parts, indicating that the segmentation results may contain outliers from the static region). Finally, by combining the point cloud sequence $\mathcal{P}c$ and the motion part segmentation masks $\mathcal{M}$, we obtain the point cloud sequence of the static region and motion parts $\mathcal{P}=\{\{P_i^{k}\}_{k=0}^K\}_{i=1}^N$.

\subsection{Initialization} 
\label{sec:Initialization} 
\noindent\textbf{Scene’s 2D Gaussian Initialization.} 
Using the first frame points cloud $\{{P_1^{k}}\}_{k=0}^K$, we initialize the scene's 2D Gaussian sets $\mathcal{GS}=\{GS^{k}\}_{k=0}^K$, where $GS^0$ represents the static region, and $GS^{k}(k\in[1,K])$ represents the segmented motion part. For ease of expression, in the subsequent description, $GS^{k}$ denotes the Gaussians of the segmented motion part. Each Gaussian's position matches its corresponding point, with its $uv$ plane normal aligned with the point's normal. Opacity, spherical harmonic coefficients, and scale are initialized following 2D Gaussian Splatting~\cite{huang20242d}. During optimization, no Gaussians are added or removed.

\noindent\textbf{\textbf{Motion Attributes Initialization.}}
\label{sec:Motion Params Init} 
In scene, each motion part Gaussians $GS^{k}$ moves as a whole, with individual motion attributes $A^k=<t_k, p_k>$.
To handle unknown motion types during initialization, we set each motion part as \textbf{RT} type, thus, the motion parameters $p_k$ include: axis position $q_k$ and direction $\vec{r}_k$.

In Section \ref{sec:Video Frames Preprocessing}, we have obtained the point cloud sequence $\{P_{i}^k\}_{i=1}^N$ for each motion part Gaussians $GS^{k}$. However, due to errors in geometric and spatiotemporal alignment of depth estimation results, only when the movement is significant, the motion pattern can be roughly reflected in the transformation between two frame point clouds. Therefore, we use the two point clouds from $\{P_{i}^k\}_{i=1}^N$ that exhibit the largest motion magnitude to initialize the motion axis, performing point cloud registration to derive the transformation matrix between them:
\vspace{-5pt}
\begin{equation}
Matrix=\begin{bmatrix}
    R & t \\
    0^T & 1
\end{bmatrix}
\label{eq:Rot}
\end{equation}

Then, we extract the axis direction $\vec{r}^{\bm{'}}$, position $q^{\bm{'}}$, and rotation angle $\theta$ from the rotational component $R$. If the rotation angle $\theta$ is greater than the threshold $\theta_{min}$, the motion parameters are initialized as $\vec{r}_k=\vec{r}^{\bm{'}}$ and $q_k=q^{\bm{'}}$. Otherwise, if the rotation angle is less than the threshold $\theta_{min}$, we consider the motion type to be more inclined towards translation. In this case, we use the translational component $t$ to initialize the motion axis direction $\vec{r}_k=\frac{t}{\|t\|_{2}}$, and set the axis position $q_k=\frac{P_1^k}{|P_1^k|}$ as the center of the first frame point cloud of the motion part.

\begin{table*}[ht]
\centering
\renewcommand{\arraystretch}{1.5}
\resizebox{\textwidth}{!}{ 
\begin{tabular}{ccccccccccc}
\Xhline{1.2pt}
 &  & \multicolumn{5}{c}{rotation} & \multicolumn{2}{c}{translation} & rotation-translation\\ \cline{3-7} \cline{8-9} \cline{10-10} 
Metrics                                       & Methods & Fridge*3 & Door & Cupboard*4 & Faucet & Laptop & Drawer*6 & Flatdoor & Liftchair & Mean \\ \hline
\multirow{5}{*}{AE(\degree)$\downarrow$} & PARIS-scene &21.252&20.204&44.946&76.275&48.311&39.633&17.714&-&37.873\\
                                              & PARIS-obj &5.984&\textbf{1.060}&17.551&22.182&87.371&6.286&52.519&-&17.000\\
                                              & DGMarbles\textsuperscript{\textbf{*}} &2.697&2.747&10.809&10.296&6.412&22.208&10.061&18.648&12.930\\
                                              & Ours(w/o optim)&2.523&2.596&6.676&12.603&5.508&24.878&6.506&6.209&12.054\\
                                              & Ours&\textbf{1.077}&1.129&\textbf{1.341}&\textbf{0.298}&\textbf{1.68}1&\textbf{1.183}&\textbf{0.280}&\textbf{1.506}&\textbf{1.262}\\ \hline
\multirow{5}{*}{PE($cm$)$\downarrow$}    &PARIS-scene&125.037&1.148&68.490&5.465&5.364&-&-&-&66.105\\
                                              &PARIS-obj&77.156&10.937&64.393&20.193&8.379&-&-&-&52.855\\
                                              &DGMarbles\textsuperscript{\textbf{*}}&19.639&2.4165&43.718&2.270&10.581&-&-&18.892&26.336\\
                                              &Ours(w/o optim)&13.456&\textbf{2.670}&14.796&2.307&8.840&-&-&9.899&11.206\\
                                              &Ours&\textbf{4.938}&11.225&\textbf{5.547}&\textbf{0.057}&\textbf{2.889}&-&-&\textbf{2.093}&\textbf{4.843}\\ \Xhline{1.2pt}
\end{tabular}
}
\caption{Quantitative results for motion axis prediction. 'Fridge*3' indicates that the fridge category contains three motion parts, with similar notation applied to other categories. Our method achieved the best results across almost all data, demonstrating a significant advantage over other methods.}
\label{tab:Comparative experiment}
\vspace{-8pt}
\end{table*}

\subsection{Optimization}
\label{sec:Optimization}
We employ the dynamic scene optimization method to identify and preserve the actual motion parts and refine their corresponding motion attributes in a zero-shot manner.

In the video, $I=\{I_i\in{\mathbb{R}^{H\times W\times 3}}\}_{i=1}^N$, each frame corresponds to a state of the dynamic scene. Between adjacent states, each motion part Gaussians $GS^k$ may undergo movement. For translation, the movement is described by the translational distance along axis between adjacent frames: $\bm{\Delta\alpha}^k=\{\Delta\alpha_{i\rightarrow(i+1)}^k\}_{i=1}^{N-1}$. For rotation, the movement is represented by the rotational angle around axis between adjacent frames: $\bm{\Delta\phi}^k=\{\Delta\phi_{i\rightarrow(i+1)}^k\}_{i=1}^{N-1}$. It should be noted that $GS^k$, axis parameters(direction $\vec{r}_k$ and position $q_k$), $\bm{\Delta\alpha}^k$ and $\bm{\Delta\phi}^k$ are all optimizable. 

In each training iteration, we randomly sample two states $A, B \in [1, N]$ and transform each motion part Gaussians $GS^k$ to the states based on the motion attribute $A^k=<t_k, p_k>$ and movement quantities. Taking to state A for example: $\Delta\alpha_{1\rightarrow A}^k = \sum_{i=1}^{A-1}\Delta\alpha_{i\rightarrow(i+1)}^k$ denotes translational distance, $\Delta\phi_{1\rightarrow A}^k = \sum_{i=1}^{A-1}\Delta\phi_{i\rightarrow(i+1)}^k$ denotes rotational angle. If the motion type $t_k=\bm{T}$, motion part Gaussians $GS^k$ will translate $\Delta\alpha_{1\rightarrow A}^k$ along axis; if $t_k=\bm{R}$, $GS^k$ will rotate $\Delta\phi_{1\rightarrow A}^k$ around axis; if $t_k=\bm{RT}$, $GS^k$ will rotate $\Delta\phi_{1\rightarrow A}^k$ around axis then translate $\Delta\alpha_{1\rightarrow A}^k$ along axis. In the implementation, these motion processes from initial state to state A can be represented by a transformation matrix $Mat_A^k$; applying it to  the position $\bm{x}$ and rotation $\bm{r}$ of $GS^k$ yields the motion part Gaussians $GS^k_A$ in state A.

Then, the scene of state A is constructed by combining the static region Gaussians $GS^0$ with all motion parts Gaussians $\{GS^k_A\}_{k=1}^K$. Similarly, the scene of state B is composed of $\{GS^k_B\}_{k=1}^K$ and $GS^0$. The scenes of state $A$ and $B$ are optimized through rendering and normal losses; meanwhile, the dynamics of each motion part Gaussians from $A$ to $B$ are optimized using the motion loss (see Sec.\ref{sec:Constraint}).

Since \textbf{T} and \textbf{R} motion types are special cases of \textbf{RT}, we assume that the motion type of each motion part Gaussians $GS^k$ is \textbf{RT} at initialization. Once a certain number of iterations $iter_{judge}$, we determine the motion type of each $GS^k$ based on the total translational movement $\bm{\alpha}^k=\sum_{i=1}^{N-1}\Delta\alpha_{i\rightarrow(i+1)}^k$ and the total rotational movement $\bm{\phi}^k=\sum_{i=1}^{N-1}\Delta\phi_{i\rightarrow(i+1)}^k$:
\vspace{-5pt}
\begin{equation}
\begin{cases} 
\bm{T} & \text{if } \bm{\alpha}^k\geq\alpha_{min},\bm{t}^k<t_{min}\\
\bm{R} & \text{if } \bm{\alpha}^k<\alpha_{min},\bm{t}^k\geq t_{min}\\
\bm{RT} & \text{if } \bm{\alpha}^k\geq\alpha_{min},\bm{t}^k\geq t_{min}\\
\bm{not\,motion\,part} & otherwise
\end{cases}
\end{equation}
Meanwhile, these Gaussians that belong to the static region but are misclassified as motion parts are pruned and merged into the static region $GS^0$. Finally, each retained (unpruned) motion part Gaussian $GS^x$ is mapped to a motion part $M^x$ with refined motion attributes $A^x$.  

\begin{figure*}[t]
  \centering
  \includegraphics[width=1\linewidth]{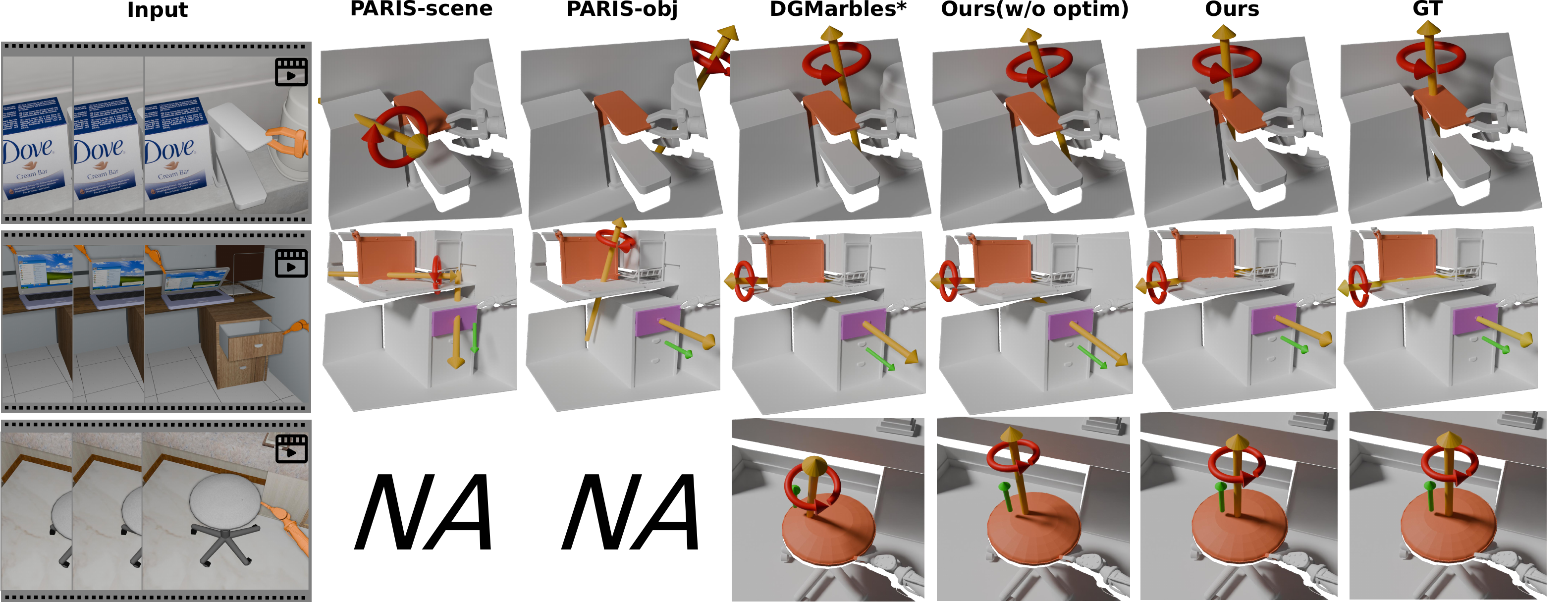}

   \caption{Qualitative results for motion axis prediction. Gray areas represent the background regions, while other colors indicate various motion parts. Red circular arrows denote rotation direction, and green linear arrows indicate translation direction. Blank image marked with \textbf{NA} denotes that the method is unable to handle this motion type. All the motion parts segmentation results are from ours, with a focus on evaluating the motion axis predictions. The comparison clearly shows that our method produces results closely aligned with the ground truth and outperforms other methods.}
   \label{fig:gallery}
   \vspace{-10pt}
\end{figure*}

\subsection{Constraints}
\label{sec:Constraint}
We formulate constraint losses from three key perspectives: rendering loss (visual consistency of dynamic scenes), normal loss (geometric consistency), and motion loss (motion pattern consistency).

\noindent\textbf{\textbf{Rendering Loss.}}
After transforming the 2D Gaussians to the corresponding state, the RGB image $\hat{I}$ is rendered, and the rendering loss is computed through comparison with the ground truth RGB image. The rendering loss is consistent with the RGB reconstruction loss used in 2D Gaussian splatting~\cite{huang20242d}, which includes $L_1$ loss and $D-SSIM$ term.
\vspace{-5pt}
\begin{equation}
L_{rend} = (1-\lambda_{dssim})L_1 + \lambda_{dssim}L_{dssim}
\label{eq:rend loss}
\end{equation}

\noindent\textbf{\textbf{Normal Loss.}}
In addition to RGB image, the Gaussians normal map $\hat{O}\in \mathbb{R}^{H\times W \times 3}$ is rendered and compared with the normal map $O$ derived from the estimation depth map:
\vspace{-5pt}
\begin{equation}
L_{normal}=\frac{\|1-|\hat{O}*O|\|_1}{HW}
\end{equation}
where, $\hat{O}*O$ denotes the element-wise dot product of the normals in the normal maps, $|\cdot|$ indicates the absolute value of the elements, $H$ and $W$ represent the image size.

\noindent\textbf{\textbf{Motion Loss.}}
Due to the depth map misalignment across frames, it is hard to accurately guide motion optimization. Therefore, this loss is only designed for the consistency of the motion pattern (particularly when the normal and rendering losses provide insufficient supervision in low-texture region and highly symmetrical motion parts), ensuring the motion of Gaussians between two frames maintains a consistent trend with the motion between point clouds.

Taking the dynamics from state $A$ to state $B$ as an example. For each motion part Gaussians $GS^k$, it is transformed from the initial state to the states of $A$ and $B$ through transformation matrices $Mat^k_A$ and $Mat^k_B$, obtaining $GS^k_A$ and $GS^k_B$ (see Sec.\ref{sec:Optimization}). Through video preprocessing(see Sec.\ref{sec:Video Frames Preprocessing}), the corresponding point cloud sequence $\{P_i^{k}\}_{i=1}^N$ of motion part Gaussians $GS^k$ is obtained. Then, we could obtain the corresponding point clouds $P_{A}^k$ and $P_{B}^k$ of states $A$ and $B$.

Our goal is to make the Gaussians motion pattern from state $A$ to $B$($GS^k_A\rightarrow GS^k_B$ ) consistent with the point cloud motion ${P_{A}^k}\rightarrow{P_{B}^k}$. Maintaining the motion pattern consistency between $GS^k_A\rightarrow GS^k_B$ and ${P_{A}^k}\rightarrow{P_{B}^k}$ means that the Gaussians' motion transformation matrix ${Mat}^k_{{A}\rightarrow{B}}=Mat^k_B {Mat^k_A}^{-1}$ 
should be roughly consistent with the transformation between point cloud $P_{A}^k$ and $P_{B}^k$. Therefore, we designed the following loss function:
\vspace{-6pt}
\begin{equation}
L_{motion} = \sum_{k=1}^K\frac{\|Mat^k_{{A}\rightarrow{B}}{P_{A}^k}-{P_{B}^k}\|_2}{|{P_{A}^k}|}
\end{equation}
where, $|\cdot|$ denotes the number of points in the point cloud.

%% file: sec/4_experiment.tex
\section{Experiment}
\label{sec:Experiment}
\subsection{Dataset} 
\label{sec:data}
To verify the robustness and broad applicability of our method, we constructed a comprehensive dataset including 15 simulated scenarios and 11 real-world scenarios, which contain various articulated objects such as drawers, staplers, elevators, etc. More details are available in the supplementary material. We collected 3D models from the 3D Warehouse~\cite{3dwarehouse2024} and built virtual scenes using Blender~\cite{blender2024}. To quantitatively evaluate the performance of methods, in addition to monocular videos, we also captured 3D point clouds with motion part segmentation and corresponding motion attribute annotations. For real-world scenarios, we recorded motion videos of articulated objects using mobile phones. Due to the lack of geometric and annotation information in real-world data, only qualitative evaluation can be performed. The experimental results (See Fig.\ref{fig:real_world}) show that our method can be effectively applied to real-world and virtual scenario datas.

\subsection{Metrics}
\label{sec:metrics}
We use the Intersection over Union ($IOU$) to evaluate the identification results of motion parts. To assess motion attributes, similar to Shape2Motion~\cite{wang2019shape2motion}, we measure the motion type accuracy ($TA$), angle error ($AE$), and position error ($PE$) between the estimated axes and ground truth. $AE$ represents the angle between the predicted motion axis and the ground truth motion axis. $PE$ represents the minimum distance between the  motion axis and the ground truth (only measured when the motion type is \textbf{R} or \textbf{RT}).

\begin{figure*}[t]
  \centering
  \includegraphics[width=1\linewidth]{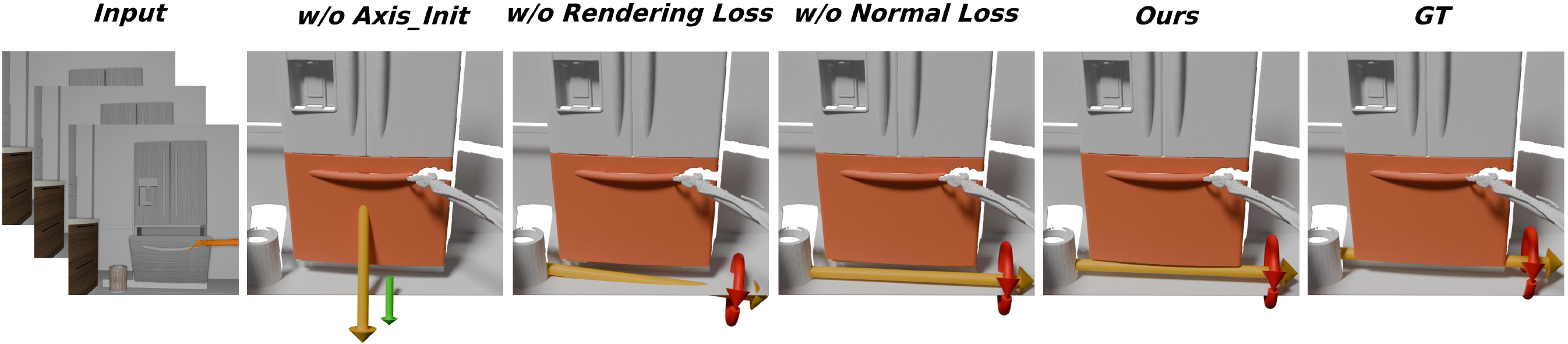}

   \caption{Ablation of our axis initialization and losses. Removing the motion axis initialization module, the judgement of motion type is incorrect. Without either the rendering loss or normal loss, the precision of the predicted motion axis decreases. Our full method achieves result closest to the ground truth.}
   \label{fig:ablation_1}
   \vspace{-8pt}
\end{figure*}

\subsection{Comparison Methods}
\label{sec:Baselines}

\noindent\textbf{\textbf{PARIS-scene and PARIS-object.}} 
PARIS \cite{liu2023paris} reconstructs the static region and motion parts of articulated objects from multi-view images of the initial and final states, and analyzes the motion attributes. However, it does not support monocular video inputs. To address this, we collected multi-view data to meet the input requirements. Additionally, since PARIS processes only one motion part at a time, whereas our scenes often feature multiple motion parts, we adapt the setup to ensure singular part movement in a scene. This adapted setup is termed PARIS-scene, with "scene" signifying that the input encompasses both the target articulated object and background context.
PARIS\cite{liu2023paris} is originally designed for object-level task. Therefore, we also designed an experimental setup by removing the background information from the scene (retaining only the target articulated object), with all other settings consistent with PARIS-scene. We refer to this as PARIS-obj. The multi-view test data examples are provided in the supplementary material.

\noindent\textbf{\textbf{DGMarbles\textsuperscript{\textbf{*}}.}}
    DGMarbles~\cite{stearns2024dynamic} uses 3D Gaussians~\cite{kerbl20233d} as the scene representation to reconstruct dynamic scenes from monocular video, dividing the scene into foreground (dynamic region) and background (static region). Each Gaussian point in the foreground moves over time. This method does not have the capability to analyze the motion parts and motion attributes. Therefore, we apply additional post-processing operations to DGMarbles, denoted as DGMarbles\textsuperscript{\textbf{*}}. Specifically, we export the positions of all foreground Gaussian points in the first frame and the last frame as point clouds and partition them into motion parts using ground truth segmentation. Subsequently, using our initialization method (see Sec.\ref{sec:Initialization}), we derive all motion parts with motion attributes.

\begin{figure}[t]
  \centering
  \includegraphics[width=1\linewidth]{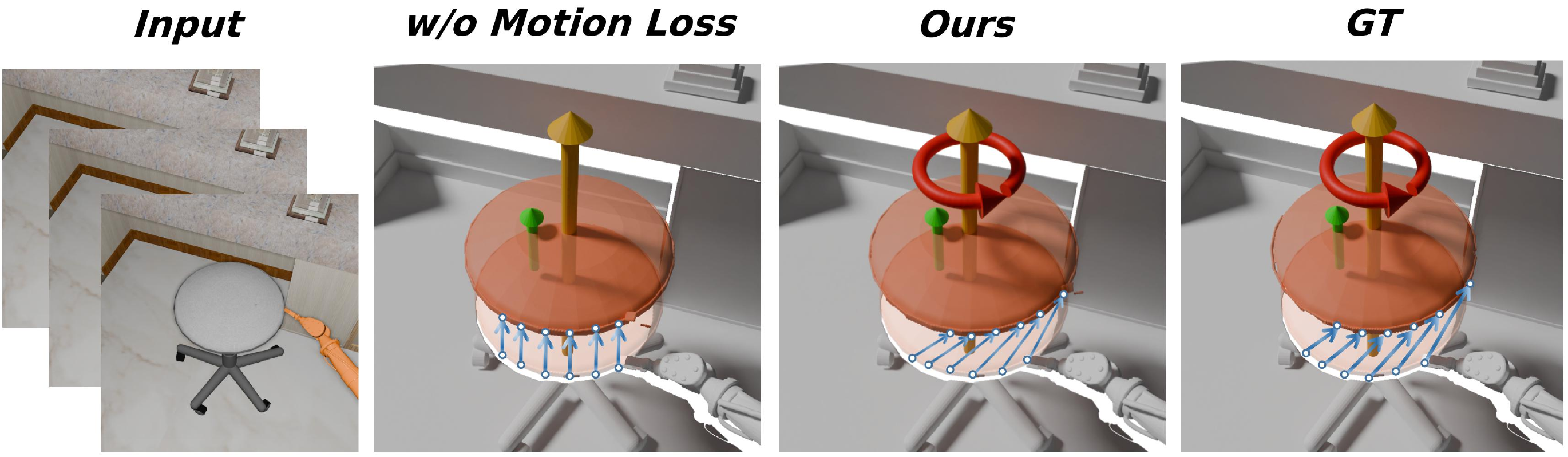}
   \caption{Ablation of Motion Loss. Without Motion Loss, the motion of the liftchair is interpreted as simple upward movement. With Motion Loss, the helical motion pattern is correctly captured.}
   \label{fig:ablation_2}
   \vspace{-8pt}
\end{figure}

\subsection{Experiment and Evaluation Setup}
During motion attribute initialization, the minimum angular threshold is set as $\theta_{min}=10\degree$. The translational distances $\bm{\Delta\alpha}^k$ and rotational angles $\bm{\Delta\phi}^k$ are initialized to zero vectors. The motion type classification thresholds are set set as follows: $iter_{judge}$=2000$, 
 \alpha_{min}=0.1*radius$ (where $radius$ denotes the radius of the smallest enclosing sphere for the motion part) and $\phi_{min}=0.05\pi$. The optimization process executes 7500 iterations with loss function weights set as: $\lambda_{dssim}=0.2$, $\lambda_{rend}=500$, $\lambda_{motion}=10$, $\lambda_{normal}=500$.

\subsection{Experiment Results}
\label{sec:Experiment Results}
PARIS-scene, PARIS-obj, and DGMarbles\textsuperscript{\textbf{*}} predict motion parameters given motion types and are unable to identify the motion parts from monocular videos. Therefore, we primarily evaluate $AE$ and $PE$. In our dataset, each scene may contain multiple motion parts with different motion types. To enable more intuitive evaluation of the methods' performance, we categorize all motion parts in the dataset by motion type and object class. The quantitative results are presented in Tab.\ref{tab:Comparative experiment}. Quantitative comparisons show that, compared to DGMarbles\textsuperscript{\textbf{*}}, our method significantly improves the precision of motion axis parameters. Due to the limited field of view, both PARIS-scene and PARIS-obj fail to capture the objects comprehensively, resulting in poor performance. Fig.\ref{fig:gallery} qualitatively demonstrates the advantages of our method in predicting motion axis. More experiment results are provided in the supplementary material.

\begin{table}[h]
    \centering
    \renewcommand{\arraystretch}{1}
    \begin{tabular}{ccccc}
        \toprule
        Setting & IOU$\uparrow$ & TA$\uparrow$ & AE($^\circ$)$\downarrow$ & PE($cm$)$\downarrow$ \\
        \midrule
        w/o $Axis\_Init$  & 0.956&0.667&10.235&8.978\\
        Ours &0.956&\textbf{1.0}&\textbf{1.143}&\textbf{4.843}\\
        \bottomrule
    \end{tabular}
    \caption{Ablation of the Motion Axis Initialization. With the Motion Axis Initialization, our method demonstrates enhanced accuracy in motion type judgement and higher precision in motion axis prediction.}
    \label{tab:ParamsInit}
    \vspace{-8pt}
\end{table}

\begin{table}[h]

    \centering
    \renewcommand{\arraystretch}{1}
    \begin{tabular}{ccccc}
        \toprule
        Setting & IOU$\uparrow$ & TA$\uparrow$ & AE($^\circ$)$\downarrow$ & PE($cm$)$\downarrow$ \\
        \midrule
        w/o $L_{rend}$  & 0.956&1.0&7.264&10.474\\
        w/o $L_{normal}$ & 0.956&1.0&4.427&6.058\\
        w/o $L_{motion}$ & 0.956&0.889&4.995&7.772\\
        Ours &0.956&\textbf{1.0}&\textbf{1.143}&\textbf{4.843}\\
        \bottomrule
    \end{tabular}
    \caption{Ablation of losses. The precision of the motion axis improves with the use of rendering and normal losses, the accuracy of motion type judgement improves with the use of motion loss.}
    \label{tab:constrains}
    \vspace{-12pt}
\end{table}

\vspace{-3pt}
\subsection{Ablation Study}
\label{sec:Ablation Study}
\noindent\textbf{\textbf{Motion Axis Initialization.}}
To validate the efficacy of our motion axis initialization strategy, we replaced it with random initialization. The experimental results in Table \ref{tab:ParamsInit} show that our initialization strategy sets the axis parameters close to the optimal values, significantly stabilizes the judgment of motion types, and effectively improves the accuracy of axis parameters prediction.

\noindent\textbf{\textbf{Constraints.}}
We use rendering loss, normal loss, and motion loss to constrain the optimization process. To assess the impact of each loss function, we perform ablation studies by individually removing each component. Table \ref{tab:constrains} shows the results, revealing that rendering loss and normal loss improve the accuracy of motion axis prediction (see Figure \ref{fig:ablation_1}), motion loss guides axis optimization aligned with the realistic motion pattern (see Figure \ref{fig:ablation_2}).

\noindent\textbf{\textbf{Ours(w/o optimization).}}
To evaluate the efficacy of the optimization module, we disabled this module and directly estimated the motion parameters of the motion parts using our initialization method (see Sec.\ref{sec:Initialization}). Specifically, given the ground-truth motion parts and the motion type , we derive motion parameters based on Equation \ref{eq:Rot}. For \textbf{T} motion type, we use the component $t$ of $Matrix$ to calculate the axis direction $\vec{r}=\frac{t}{\|t\|_{2}}$. For \textbf{R} or \textbf{RT} motion type, we derive the axis direction $\vec{r}$ and position $q$ from the component $R$ of $Matrix$. The quantitative experimental results  (See Tab. \ref{tab:Comparative experiment}) show that the optimization module can significantly improve the estimation results of motion parameters.

%% file: sec/5_conclusion.tex
\vspace{-3pt}
\section{Conclusion}
\label{sec:Conclusion}
\vspace{-5pt}
This paper presents \textit{MonoMobility}, a novel zero-shot framework designed for parsing the motion parts and their motion attributes (including motion axes and motion types) of articulated objects from monocular videos. The core of the framework is a self-supervised dynamic scene analysis method specially designed for articulated objects, which mainly consists of coarse estimation stage and fine optimization stage. Specifically, during the coarse estimation stage, we leverage existing mature depth and optical flow estimation methods, combined with the inherent characteristics of the video stream, to obtain prior information on scene geometry and motion, thereby enabling preliminary parsing of motion parts and their attributes in the scene. In the fine optimization stage, we integrate 2D Gaussian splatting and introduce additional normal and motion loss regularization terms to jointly optimize the motion parts and their attributes. Furthermore, to validate the effectiveness of our method, we constructed a dataset for articulated object motion estimation, encompassing a wide range of synthetic and real-world scenes, with objects covering common articulated motion types encountered in daily life. Extensive qualitative and quantitative experimental results demonstrate that the proposed method significantly outperforms baseline methods.

However, our method still has some limitations. First, similar to other dynamic scene reconstruction techniques, our approach requires individual optimization for each scene, which may limit its application scenarios and scope due to efficiency concerns. Second, our method relies on offline depth estimation and optical flow estimation methods. Although these methods have achieved near-maturity in terms of efficiency and quality, they may perform poorly in certain special scenarios, leading to failures in the initial parsing results and thereby restricting the framework's applicability. More detailed analysis can be found in the supplementary material. We regard these challenges as future research directions to further improve the robustness and practicality of the method. 

%% file: sec/7_acknowledge.tex
\vspace{-3pt}
\section{Acknowledge}
\label{sec:Conclusion}
\vspace{-5pt}
This work was supported in part by the NSFC (62325211, 62132021), the Major Program of Xiangjiang Laboratory (23XJ01009), the Chongqing Natural Science Foundation (CSTB2024NSCQ-MSX1026), the Fundamental Research Funds for the Central Universities (No. SWU-KT25012), Key R\&D Program of Wuhan (2024060702030143).